# Grading and Anomaly Detection for Automated Retinal Image Analysis using Deep Learning

Syed Mohd Faisal Malik [1] , Md Tabrez Nafis [2*] , Mohd Abdul Ahad [3],

Safdar Tanweer [4]

[1]Ph.D Research Scholar, Jamia Hamdard,India

[2,3,4]Associate Professor,Jamia Hamdard,India

Corresponding author: email: Tabrez.nafis@gmail.com

Emails of Coauthors:

[1]Fslmalik9@gmail.com;[3]itsmeahad@gmail.com;[4]safdartanweer@yahoo.com

## Abstract

The significant portion of diabetic patients was affected due to major blindness caused by Diabetic retinopathy (DR). For diabetic retinopathy, lesion segmentation, and detection the comprehensive examination is delved into the deep learning techniques application. The study conducted a systematic literature review using the PRISMA analysis and 62 articles has been investigated in the research. By including CNN-based models for DR grading, and feature fusion several deep-learning methodologies are explored during the study. For enhancing effectiveness in classification accuracy and robustness the data augmentation and ensemble learning strategies are scrutinized. By demonstrating the superior performance compared to individual models the efficacy of ensemble learning methods is investigated. The potential ensemble approaches in DR diagnosis are shown by the integration of multiple pre-trained networks with custom classifiers that yield high specificity. The diverse deep-learning techniques that are employed for detecting DR lesions are discussed within the diabetic retinopathy lesions segmentation and detection section. By emphasizing the requirement for continued research and integration into clinical practice deep learning shows promise for personalized healthcare and early detection of diabetics.

## 1. Introduction

"Diabetic Retinopathy (DR)" is specified through progressive vascular disruptions in the retina. This disruption is developed by the patient with diabetes and is created by chronic hyperglycaemia based on its severity. Among the adults' work nature, has been considered the major cause of blindness worldwide. Among diabetic patients, the study proved that globally, there are almost 27% of the people have DR [1]. It is expected that these numbers can be increased more due to the elevating pervasiveness of diabetics in Asian countries like China and India [2]. In the primary stages, although DR is totally asymptomatic. During these primary stages, clinically invisible microvascular transformation and neural retinal damage progress strongly[3]. Timely solutions and managing of disease in an efficient way was achieved by proceeding with regular eye screening [4]. Furthermore, to avoid this problem, the only way is to control hypertension, hyperlipidaemia, and hyperglycaemia, and DR's early detection has become necessary [3]. Regarding its treatment recently, there have been several available interventions, including laser photocoagulation. Suppose at the disease's primary stage the eyes are checked. In that case, this intervention remarkably decreases the blindness chances in diabetic maculopathy and proliferative retinopathy to 98% [5]. Finally, it has been proved that from diabetic retinopathy, blindness prevention is possible through appropriate treatment and early detection[6].

DR's initial diagnosis depends on functional transformation in "retinal blood vessel caliber", "retinal blood flow", and "electroretinography (ERG)"[7]. Early diagnosis relies on fundus evaluation in clinical practice [8]. Kwan and Fawzi[9], highlighted that fundus photography is considered a widely available, well-tolerated, non-invasive, and rapid imaging technique. It comprises a major applied technique to analyse the DR's extension. To identify DR and to predicate effects Ophthalmologists observe the lesions of the retina by applying fundus images at high resolution. In Highly populated countries like India and Africa, people residing in rural area need of images diagnosing DR manually demands a strong effort and expertise level by a professional ophthalmologist. In these countries, identification of diabetic problem, the number of individuals having DR and diabetes will enhanced dramatically, and there will be a disproportionately low number of ophthalmologists[10]. The research community is motivated by this contradiction and decided to introduce computer-aided diagnosis that assists in reducing effort, time and cost required by an expert medical technician to detect DR. Apart from that, computational capabilities and resources enhancement and recent shifts in "Artificial Intelligence (AI)" have created a scope to launch applications integrated with "Deep Learning (DL)" associated with appropriate identification and types of DR. During previous few years, certain articles have been published that discussed on the methods of DL and its impact on DR[11–15]. Maximum articles emphasized the particular perspectives of the modelling pipeline and data analysis.

In some cases, it is restricted to the model's performance reporting to the pre-processing techniques that are commonly used[13,14]. The researcher indicated data sets that are publicly available, but complete description of information was not available. Therefore, a more integrated and detailed method has been implemented for these fragmented efforts to evaluate application area of the study integrated with technical data was considered for evaluation. The study shows a complete viewpoint regarding "analysis pipeline". Furthermore, besides illustrating comparative technical data was available in the internet regarding DL models' development for the segmentation and classification images related to fundus, the study additionally focused on discussing publicly available data sets' thorough analysis, the commonly applied pre-processing pipelines, and in the real setting of the clinic, as a models' presentation these have been used.

## 2. Methodology

The proposed review article significantly follows the significant principles of systematic literature review concerning the PRISMA guidelines. The PRISMA guidelines exploring the process of identifying the articles, screening of suitable documents, eligibility and inclusion of opted documents. The study successfully conducted systematic review through the Scopus data base, and the articles are mainly published in "IEEE, Wiley, Springer, Hindawi, MDPI, and Science direct". The documents are collected over based on basic building blocks of words "Deep Learning and Diabetic Retinopathy". The investigation results stating that over 38.8% full-text research articles, 47.3% of conference papers, 5.6% of conference review articles, and the rest of publications are from other categories (book chapters, review article, editorial, books, note, retracted, and erratum).

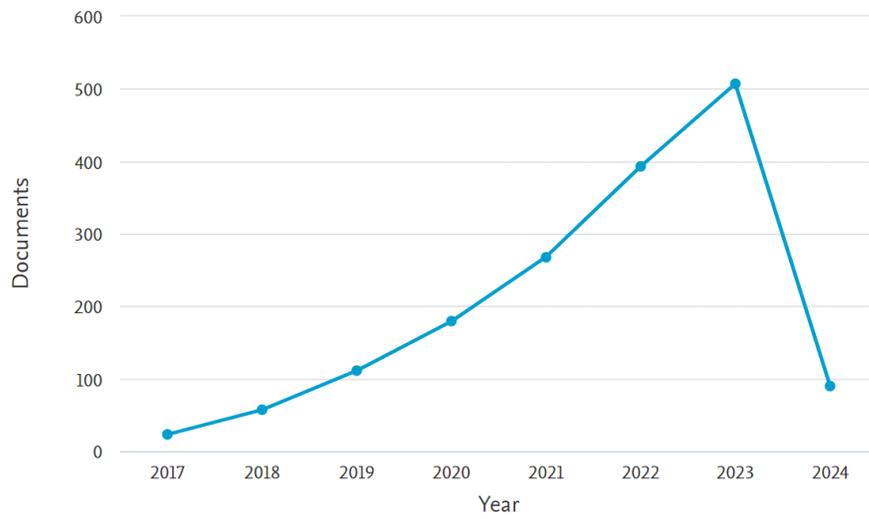

Figure 1: Documents published over 2017 to 2024 in Scopus

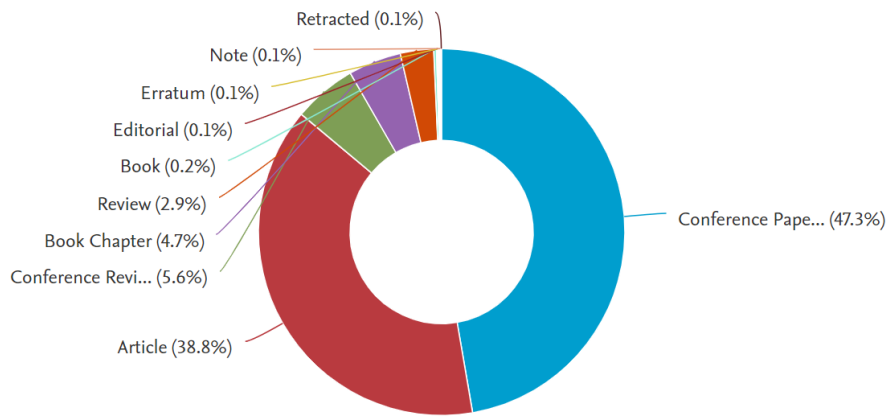

Figure 2: Documents classification over 2017 to 2024 in Scopus

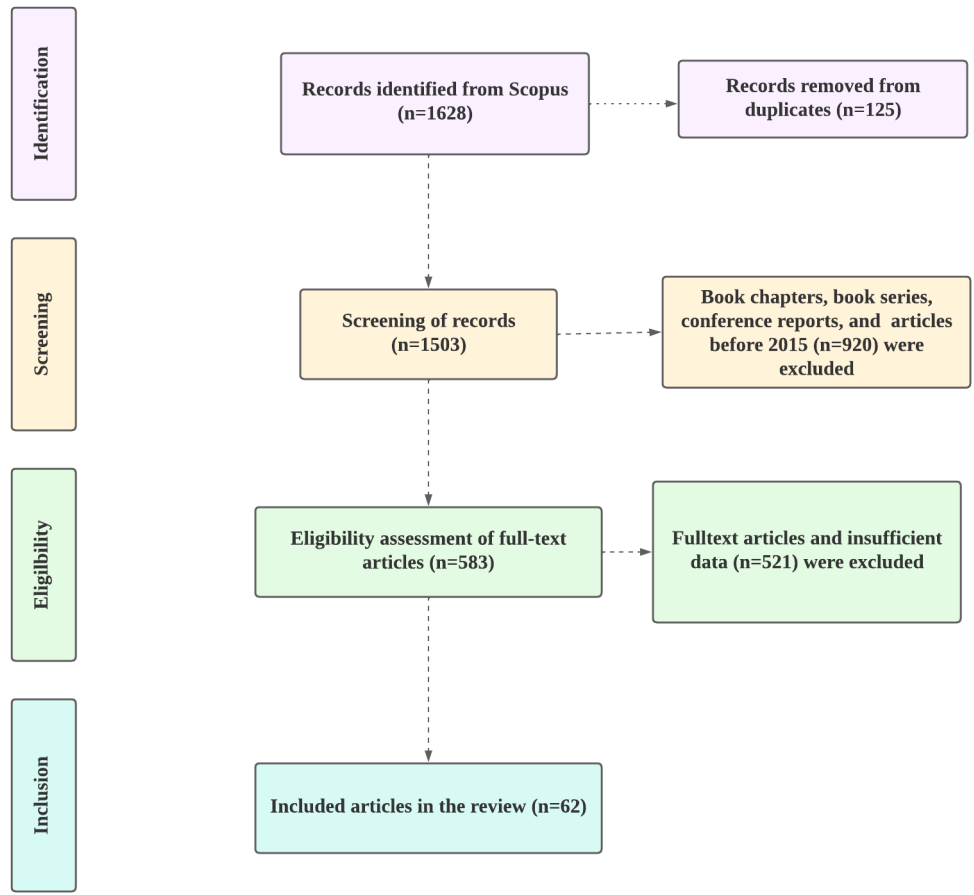

Figure 3: PRISMA framework

The PRISMA model of the current systematic literature review was predicated by using the diagram3. The PRISMA follows "identification, screening, eligibility, and inclusion". In the identification criteria, the overall records identified from the database is 1628 documents, from the 1628 documents, 125 records have been removed concerning the duplicates and 1503 articles has been processed into the screening section. In the screening of records, 1503 articles have been included, from the documents, 920 articles have been excluded concerning the book chapters, book series, conference reports, and erratum. In the eligibility criteria, the overall studies processed are 583 documents, and 521 articles have been excluded concerning the full-text articles and insufficient criteria. And finally, the review included 62 documents relating towards the cloud broker in the present research.

## 3. Diabetic Retinopathy

Microaneurysms have been noticed on the retina during diabetic retinopathy's primary stages and result from loss of pericytes and degeneration that create capillary wall dilatation. Intraretinal hemorrhages occur when the microaneurysm's or capillary's wall is ruptured. There are some other lesions of "non-proliferative diabetic retinopathy" like reduplication or venous loops, venous beading, "intraretinal microvascular abnormalities (IRMA)", and hard and soft exudates[16]. Tsiknakis[17] reported that huge "caliber tortuous vessels", IRMAs appear based on ischemia parts and indicate attempted vascular remodelling. Moreover, on the neovascularization's presence, the difference between non-proliferative and proliferative therapy based on detection of DR is dependent and significantly indicates the new retina vessel growth because of ischemia to previous ones.

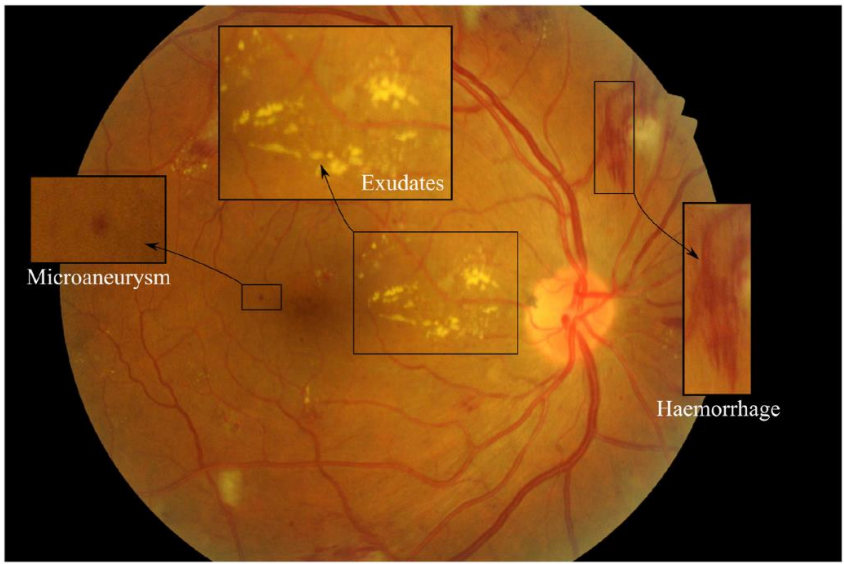

Figure 4: Diabetic Retinopathy Lesions[17]

"Diabetic macular edema (DME)"disease can be identified at any stage was stated[18]. It is contemplated as an endpoint for blindness. Apart from that, through abnormalities including the haemorrhages or microaneurysms of the retina within the fovea center's one-disc diameter, thickening of the retina within the fovea center's one-disc diameter, exudates within the "macula" and exudates within fovea center's one-disc diameter[19]. Regarding the DR's clinical grading protocols, "Early Treatment Diabetic Retinopathy Study (ETDRS) grading scheme" considered one of best method[20]. Basically, this gold standard's application in clinical practice regularly is not to be practical or easy. In an effort, many substitute scales have been introduced to develop communication among caregivers and screening of patients. In several countries, the development of these scales to evaluate the effectiveness of DR so far not caused "single international severity scale". Finally, "International Clinical Diabetic Retinopathy Disease Severity Scale" has been developed by "Global Diabetic Retinopathy Project Group"[21].

## 3.1 Traditional Approaches

The program of DR screening was developed for the first time in the "1980s and early 1990s"[22]. This program minimized the DR-associated vision loss cases successfully at that time. Hristova[23] conducted a review on diabetic retinopathy and indicated that through the teams of mobile fundus photography, early DR detection was initiated in Sweden to reduce the vision loss cases due to DR by almost 47 % every year was estimated with in the period of 5 years in 1990. Recently, some criteria have been provided by the "International Council of Ophthalmology (ICO)" initiates the screening DR that includes test of the retina with fundus images consisting of mydriatic fundus images or non mydriatic fundus images with "≥ 30° mono- or stereo photography" or indirect or direct ophthalmoscopy[24]. The grading of fundus images is necessary to measure the DR's severity level. To an ophthalmologist, these fundus images' careful manual grading can be subjective, time-consuming, and labor-intensive. In evaluating the fundus images of DR, an automated algorithm has a critical role. However, the algorithms that are previously applied are basically dependent on the traditional methods. These traditional techniques were involved with selecting the features of engineering images manually for some classifier techniques like random forest and support vector machine[25]. For Diabetic Retinopathy treatment artificial neural network (ANN) was proposed [26]. This technique has the power to grade DR automatically with 83% specificity and 88% sensitivity compared to an ophthalmologist. Furthermore, DL is regarded as a branch of AI that concentrates on learning task-specific features of DR images with different images without calculating the selection of manual features. Hence, in different DR issues a "convolutional neural network (CNN)" is considered as a major solution[27]. For the DR's automatic diagnosis, the application of CNN has provided improved performance.

## 3.2 Manual Grading by DR

DR-related primary care involves patients' vigilant follow-up for earlier and better treatment, screening of patient retina's high-quality photograph, and timely ophthalmic examination. Some common techniques, including "monochromatic photography technique" or "nonmydriatic or mydriatic digital color", "stereoscopic color film fundus photography", and "direct and indirect ophthalmoscopy" have been applied to classify and detect DR. For the classification and detection of DR, the popular gold standard technique is "stereoscopic color fundus photographs in 7 standard fields (300)". According to the group "Early Treatment DR Study (ETDRS)", this technique is recommended[20]. Through this method, "subtle retinal neovascularization" and "diabetic macular edema (DME)" have been identified accurately[28]. Moreover, this technique is more reproducible and accurate. Still it requires a huge skilled healthcare infrastructure like equipment of good photography processing, photograph readers, and photographers.

## 4. Deep Learning Techniques

"Deep Learning (DL)" is considered part of the methods of AI, and it relies on unnatural neural networks and is inspired by the human brain structure. DL primarily indicates mathematical representation and methods of learning the intrinsic and latent connections of the information automatically. Deep learning, unlike conventional machine learning techniques, needs less guidance from humans as they are not dependent on hand-crafted features. This hand-crafted task is very time-consuming and laborious. Whereas the scale of DL methods is much better compared to conventional methods of ML as the quantity of information is enhanced. This section has presented a brief overview of the concept of DL[29].

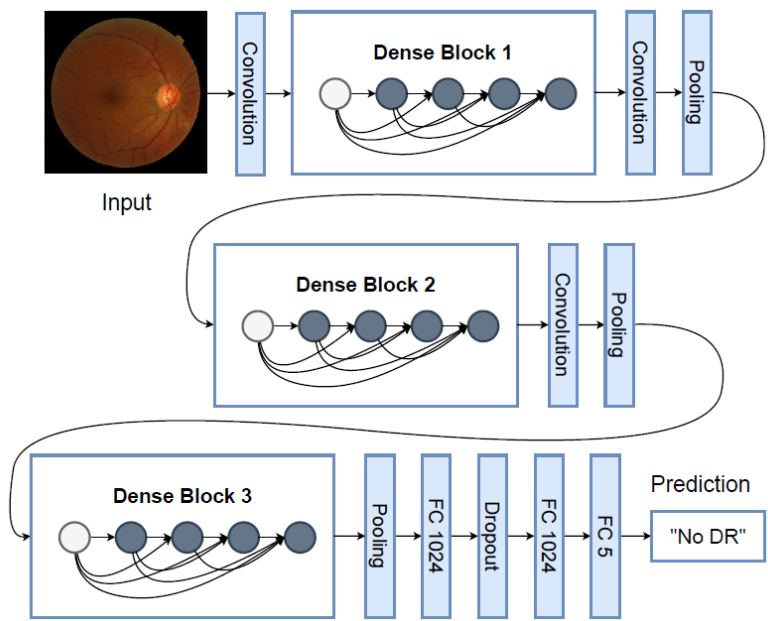

Figure 5: Diabetic Retinopathy using technique of Deep Learning[29]

In DR classification and detection, DL has been applied globally in recent years. This deep learning has several heterogeneous sources incorporated can learn appropriately the characteristics of input data [30]. Some common DL-based techniques like sparse coding, auto encoder, "convolutional neural networks (CNNs)", and restricted Boltzmann Machines[31]. When the training data's number increases, these methods' performance also increases because of the learned features' enhancement unlike the methods of machine learning [32]. In contrast, the extraction of hand-crafted features is unnecessary for DL. In the analysis of medical images, CNNs are more applied methods compared to others[33].Three major basic layer included in the architecture of CNN: "fully connected layers (FC)", convolution layers (CONV) and Pooling layers.Size,layers and CNN filters are varied according to the perspective of authors. In CNN architecture, each layer has an influential role. Different filters included in "CONV layers" impact picture to take out the characteristics. Whereas, to reduce the feature maps' dimensions, typically layer of CONV followed by the pooling layer. For pooling, there are several ways but max pooling and average pooling are mostly applied [31]. To explain the entire input image, the FC layers are perfect. However, the most applied function of classification is the SoftMax activation function. Apart from that, there are various pre-trained architectures of

CNN on the dataset of ImageNet like He[34] suggested ResNet, Szegedy[35] indicated Inception-v3 and Krizhevsky[36] presented AlexNet.The emphasized transfer learning, and the authors explored that these pre-trained architectures' transfer learning assists in speeding up the training, whereas for classification, other studies from scratch generate their own CNN[37,38]. Finally, the pre-trained models' transfer learning strategies include training the pre-trained model's all layers or finetuning multiple layers and finetuning the last "FC layer".

**4.1 Diabetic Retinopathy Retina Dataset**

For the retina, many datasets are publicly available to vessels and DR detection. Furthermore, to test, validate, and train the systems these datasets are applied frequently. Apart from that, it also helps in comparing the performance of the system against other systems. Abràmoff[39] proposed two classification of "retinal imaging", such as "optical coherence tomography (OCT)" added with "fundus color images". OCT images refer to retinas' "2 and 3-dimensional images" considered by applying low-coherence light. These images deliver critical information regarding retina structure and thickness. In contradiction, fundus images indicate retinas' "2-dimensional images" that have been collected by applying reflected light. A few years ago, "OCT retinal images" were launched. Moreover, some diversity has been observed in the data sets of fundus images that are publicly available and commonly applied. The datasets of fundus image are:

Table 1: Classification of Dataset

| S.No | Dataset | Description |
|---|---|---|
| 1 | *Kaggle* | With various resolutions, it holds Eighty eight thousand seven hundered and two "high-resolution images" that range starts "433*289 pixels to 5184*3456 pixels", captured using different type of photo capturing devices. Information on Training of images was available as open book accessible by all people. The five stages of DR, all images are divided. The collection of data includes storage of several images with incorrect labelling and poor quality[37,40,41]. |
| 2 | *DIARETDB1* | This dataset contains 89 retinal fundus images that are publicly available with "1500*1152 pixels" captured with "50-degree field of view (FOV)". Besides that, this contains five normal pictures and 84 images of DR illustrated using four skilled personnel in the medical field[42]. |
| 3 | *DDR* | It stores 13,673 fundus images that are presented to the five stages of DR and captured at a "45-degree FOV". From this dataset, 757 images are annotated to the lesions of DR[37]. |
| 4 | *E-ophtha* | It is one of the popular publicly available datasets that consists of mainly two types such as E-ophtha EX and E-ophtha MA. However, 35 normal images are included in E-optha and 47 EX-related images.Additionally, E-ophtha MA focused on containing 233 normal images and 148 MA- associated images[43]. |
| 5 | *HRF* | The dataset also provide images related to blood vessels based on segmentation. It stores 45 images with "3504*2336 pixels". Additionally, there are 15 glaucomatous images, 15 healthy images, and 15 DR images[44] |
| 6 | *Messidor* | It consists of 1200 fundus images that are colored and illustrated to the four stages of DR. The images were captured at a "45-degree FOV"[45]. |

| 7 | *DRIVE* | Dataset was applied for the segmentation based on blood vessels. This stores 40 images with "565*584 pixels" and is captured at a "45-degree FOV". Out of these 40 images, seven images are classified as mild DR and the rest are classified as normal[46]. |
|---|---|---|
| 8 | *Messidor-2* | It includes 1748 pictures captured with "45-degree FOV"[47]. |

Quellec[48] suggested the DIARETDB1 datasets for detecting the lesions of DR. Similarly, Orlando[49] identified red lesions, E-ophtha and DIARETDB1 applied, whereas Chudzik[50] to detect MA; applied these datasets. Before using the datasets for the techniques of DL, maximum studies processed these datasets properly.

### 4.2 Image Pre-processing of DR

From images to removing the sound, image pre-processing is a very important part. The consistency of the image was ensured to increase the feature of the image[51]. In this section, the researchers have discussed the most popular pre-processing methods. Several scholars set the size of the captured images with specific resolution that will appropriate suits for network applied[48,52]. To remove the images' extra spaces, cropped images were used, the pre-processing of test images are shown in the figure 6.

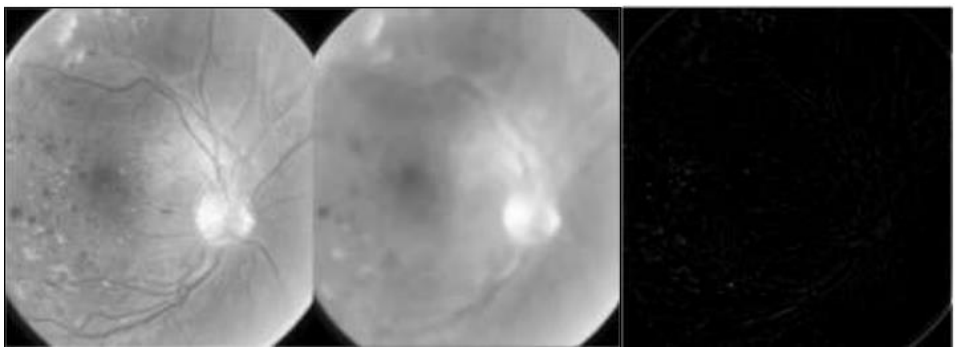

Figure 6: Pre-processing of test images[51]

In contrast, for image normalization in the same distribution, the normalization of data was applied[53]. The methods of noise removal are "NonLocal Means Denoising methods", Gaussian filter, and median filter[49,51,53]. When some classes of image were imbalanced or to enhance the size of the dataset, data argumentation methods were followed[53]. Moreover, these data argumentation techniques are rescaling, contrast scaling, flipping, shearing, rotation, and translation. For contrast enhancement, a morphological technique was applied[50]. For feature extraction, Adem[54] applied the canny edge technique. The images are prepared after pre-processing them to be applied as a DL's input.

### 5. Diabetic Retinopathy Classification

DR was categorised as two types such as identification of DR and captured screenshot related to fundus caused in retina.

### 5.1 Grading Scale

By using different classification systems, the experiments regarding diabetic retinopathy (DR) grading are conducted during the research carried [40]. Under 2-class categorising classification, the disease presence is identified. Under 3-class classification indicates not identified DR, mild DR, and severe DR. whereas under 4-class classification various types encountered are DR with severe,DR with moderated with mild and no occurrence of DR. As the presence of at least moderate non-proliferative diabetic retinopathy lesions the referable DR is defined. As the presence of at least proliferative diabetic retinopathy (PDR) lesions the vision-threatening DR is identified. Researchers capture similar clinical manifestations across various traditional scaling protocols the 4-class scale is introduced [55]. Additionally researchers conducted a study among diseased and healthy individuals to distinguish the development of a binary classification model[12]. Furthermore researchers conducted a study, for classifying the images that are ungradable is used as an additional class[37].

### 5.2 Performance Evaluation Metrics

By using a confusion matrix, the binary classification model performance is illustrated. Each entry in matrix are calculated based on the actual predication of the model. From these metrics, there are several specialized evaluation measures like Cohen's Kappa, Sensitivity, Accuracy, F1 Score, and Specificity emerged. According to curve based on receiver operating characteristic, the sensitivity of plotting against specificity across several threshold settings for classification outcome. Lastly, by measuring the area under the ROC curve the overall classifier performance is quantified by the Area Under the Curve (AUC). Across all classification thresholds, a comprehensive performance assessment is provided by ensuring this curve.

### 5.3 Binary Screening

Xu [52] noticed that by using CNN the Kaggle dataset is automatically classified into the normal images. From the dataset, nearly 1000 images can be used and before feeding pictures to CNN the augmented data and size changed to 224*224*3 performed. Based on applying different type of transformations like translation, flipping augmented data and rescaling is used in increase dataset images. Layers based on 4-max pooling, CONV layers splitted with 8 major classification, and FC layers with 2 classification are included within the architecture of CNN. For segregation, the function SoftMax is used at the CNN last layer with a value of 94.5% of accuracy. Based on another research conducted[48] it is noticed that by using three CNNs the pictures were categorised as referable DR which indicates at moderate stage with least value, or non-referable DR which indicates mild stage or no DR.

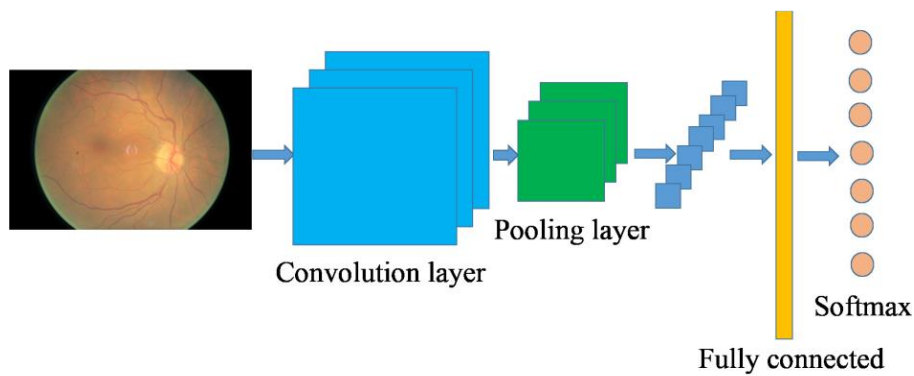

Figure 7: DCNN architecture [52]

From three datasets the images are sourced and they are identified as Kaggel, private E- ophtha, and DiaretDB1. The size of the images are changed, normalized, and minimised to pixel of 448*448. during preprocessing stage. Additionally, the augmented data is generated and the Gaussian filter is applied. The trained AlexNet is included within the employed CNN architectures and from solution the two networks are employed.microaneurysms, hard exudates,hemorrhages and soft excludates are considered as well trained CNN dataset for detection. In ROC curve, an area 0.954 is achieved considered for  Kaggle and 0.949 is achieved for datasets uses E-ophtha. According to another study conducted[56], for detecting the DR images and red lesions using 65*65 augmented patches the models of CNN were developed. By using one FC layer, five layers of CONV, and layers with five max-pooling are pretrained VGG16 is utilized. With the help of the DIARETDB1 dataset, these models were trained and for classifying patches into red lesions these models were tested on DDR, Kaggle, and Messidor. Subsequently, as non DR or DR images were categorised based on the map of test images with lesion probability. The AUC of 0.912 and the sensitivity of 0.94 is achieved for the Messidor dataset. However, it is not required to consider all the five stages of DR that causes crucial for preventing progression to blindness and determining appropriate treatment.

### 5.4 Multi-Level Classification of DR

According to another study conducted by Ayala[29], CNN method was used to detect retinopathy based on diabetic and macular edema based on diabetic. For model testing the Eyepacs-1 dataset and Messidor-2 were utilized by CNN-based method. The images underwent normalization to the width of 299 pixels and by using pretrained inception-v3 ten CNNs were effectively trained. By linear averaging functions, the final results are computed. The categories like moderate or worse DR, full gradable, referable diabetic macular edema, and severe or worse DR are included under the image classification. In both datasets and sensitives, a specificity of 93% is achieved. For Messidor-2 96.1 % is achieved and for Eyepacs-1 datasets 97.5% is achieved respectively.

However, is noticed that the five stages of DR and non-DR images were not explicitly detected. By comprising 1748 images the data augmentation is applied to the dataset based on Messidor-2. For detecting both normal retina anatomy and lesions DR the multiple CNNs are integrated with the usage of Random Forest Classifier. The vision-threatening DR, no DR, and referable DR are included as a few of the categories under the classification of images. According to the report, under the 0.980 curve, a specificity of 87.0%, and a sensitivity of 96.8%. Unfortunately, as no DR the images of mild DR stage are categorized and even five DR stages are not considered. Where another study proposed that from the Kaggle dataset, the images are classified by using CNN-based methods into five DR stages[28]. For 512*512 pixels the images are resized and the color normalization is performed during the preprocessing. The three fully connected layers, 10 convolutional layers, and the eight max-pooling layers are comprised under the custom CNN architecture. As a classifier for 80,000 test images the function of SoftMax is served. Within CNN the dropout methods and the L2 regularization are employed for controlling the overfitting. An accuracy of 75%, specificity of 95%, and sensitivity of 30% are exhibited by the obtained results.

## 6. Related Studies on Deep Learning Approaches

During the study performed one of the pioneering studies was conducted to utilize a CNN-based model for mirroring the clinical grading protocol and five-class classification of diabetic retinopathy (DR). A class-weighted strategy is used by the author for mitigating the overfitting and updating parameters during the back propagation for every batch. For enhancing the proliferative diabetic retinopathy and non-proliferative diabetic retinopathy prediction the five-class classification problem is approached by converting it into a regression task. The model performance can be boosted and the CNN feature vectors from each eye are amalgamated by the devised blending network. Similarly, a CNN model was developed[57] to analyse images from eyes and for classification fuse their representation. For maintaining the receptive field as possible to the original image size the CNN models are advocated to use network architecture adaptations and smaller convolutions.

Based on another study conducted by employing smaller filters in Conv2D layers especially detecting smaller lesions like microaneurysms the improved DR classification performance were identified[58]. For capturing lesion marks of varying sizes and capturing features at different resolutions the inception modules were employed in several studies. Whereas, how the algorithm training performance is impacted by the adjudication grading system is investigated by another author. When compared to majority voting by noting slight performance enhancement and using the adjudication grading protocol the pre-trained model is fine-tuned on the small labelled dataset.

For enhancing the detection performance the attention modules are integrated into CNNs. For enhancing classification performance on subtle regions the bilinear strategy and attention mechanism are employed by another author. The high attention regions for boosting classification accuracy are focused by aiding Crop-network and the attention maps are generated by using the attention mechanisms. Based on the author[37] exploring inter-disease correlations and detecting diabetic retinopathy a novel architecture incorporating attention modules is incorporated. The original fundus image for DR severity classification is complemented and for lesion detection, the deep learning pipeline is devised[59]. For mitigating missing lesion annotation impact the lesion clustering method is employed during the detection phase. By attention fusion network the detection model from lesion maps is fused with feature maps of the classification model. For improving the lesion segmentation performance using image-level annotated data the attention mechanism is incorporated by introducing a collaborative weakly-supervised learning model[60].

## 7. Ensemble Learning Approaches

For disease grading one ensemble model and for disease identification, another ensemble model is devised[55]. The multiple pre-trained networks are integrated by these models for feature extraction and serve as a classifier by coupling with customer standard dense neural network. The individual ones in both tasks are surpassed by ensembling these models and achieving 98.56% specificity and 98.10% sensitivity. With the strength of the base learner, the performance tends to increase and in certain cases, it is identified that the dual ensemble outperforms a single ensemble. Based on inception V3, Inception-Resnet-V2 architecture, and ResNet152 the Adaboost classifier is employed on three models by crafting an ensemble model. The ensemble model is effectively trained on a private dataset that is developed in the collaboration with Adaboost classifier[61]. By achieving 88.21% accuracy, 85.57% sensitivity, and AUC 0.946 the individual models are outperformed by the ensemble model. The superior specificity at 91.46% is exhibited notably by the InceptionV3. During the training process, the multiple iterations are exported by ensuring the training CNN model. At different training iterations, every distinct lesion type is detected optimally. By using ensemble learning these saved models are combined subsequently and help in predicting the diabetic retinopathy severity score.

## 8. Diabetic Retinopathy Lesions Segmentation and Detection

For classifying and detecting certain types of diabetic retinopathy lesions there are several approaches are outlined within this section. The red lesions in DR images are detected exclusively by utilizing deep learning methods alongside domain knowledge for feature learning[49]. After processing the images from DIARETDB1, MESSIDOR, and E-ophtha the Random Forest method for classification is employed. By applying the r-polynomial transformation, and employing multiple morphological closing functions the field of view is expanded by involving green band extraction. For convolutional neural network training the red lesion patches are augmented and resized to 32*32 pixels. The 89, 381, and 1200 images are comprised of the DIARETD, MESSIDOR datasets, and E-ophtha. By achieving competition metric scores of 0.4874 and 0.3683 the four convolutional layers, one fully connected layer, and four convolutional layers are considered within the custom CNN architecture.

For detecting the microaneurysms from DR images the custom CNN architecture and from E-ophths(381 images), DIARETDB1(89 images) and ROC (100 images) the datasets are utilized. The resizing, green plane, and cropping are extracted by involving the preprocessing and for employing morphological functions the Otsu thresholding is applied. The random transformations were applied and the MA patches were extracted. By batch normalization layer, four skip connections, three max-pooling layers, and three simple up-sampling layers in reported ROC score of 0.355 19 convolutional layers are comprised by the CNN. Whereas by integrating the enhanced pretrained LeNet architecture with a random forest classifier another study conducted and mainly focused on detecting DR red lesions in the DIARETDB1 dataset[62].

By applying the contrast-limited adaptive histogram equalization for enhancement, cropping the green channel, and removing noise with a Gaussian filter the images are enhanced by the improved pretrained LeNet architecture. By using U-net CNN architecture the blood vessels were segmented. The one fully connected layer, four convolutional layers, and three max-pooling layers are comprised of the improved LeNet architecture. By integrating handcrafted and custom CNN features with a random forest classifier lesion detection in E-ophtha datasets the hard exudate lesion detection is targeted. The candidate detection using dynamic thresholding, color normalization, and cropping the preprocessing steps are performed. For classifying and detecting DR images the remaining lesions in images after vessel extraction are essential to consider. As reviewed in this section the utilized deep learning methods are studied by a few vessel segmentation. Based on pre-trained DEEPLAB-COCO-LARGEFOV the modified CNN is employed to extract the retinal blood vessels from RGB retina images. Under ROC the area of 0.894 is achieved and 93.94% accuracy is achieved[55]. With patch cropping, horizontal flipping, and morphological methods the CHASE DB1 datasets and STARE are included under the preprocessing datasets.

## 9. Conclusion

In conclusion, the critical assessment of deep learning methodologies help in evaluating the automated identification of DR from images of retina. In the realm of diagnostics methodologies associated with deep learning and medical imaging, a significant stride has been signified. While evaluating severity and accurately discerning diabetic retinopathy the remarkable potential of deep learning techniques has been underscored by this comparative scrutiny. The challenges such as model interpretability, dataset heterogeneity, and applicability across diverse demographics the achieved progress notwithstanding. Looking ahead, the integration into clinical protocols, the validation of expansive datasets, and concerted endeavors toward standardization are imperative to fully harness deep learning in diabetic retinopathy diagnosis. Personalized healthcare, enhanced patient outcomes in diabetic retinopathy management, and the promise of bolstering early detection are held as technological advancements persist.

## 10. Acknowledgement